\documentclass[10pt,twocolumn,letterpaper]{article}

\usepackage{cvpr}              
\usepackage[accsupp]{axessibility}  

\definecolor{cvprblue}{rgb}{0.21,0.49,0.74}
\usepackage[pagebackref,breaklinks,colorlinks,allcolors=cvprblue]{hyperref}

\def\paperID{} 
\def\confName{CVPR}
\def\confYear{2025}

\title{Objaverse++: Curated 3D Object Dataset with Quality Annotations}

\author{Chendi Lin\\
Carnegie Mellon University\\
{\tt\small chendil@alumni.cmu.edu}
\and
Heshan Liu\\
Carnegie Mellon University\\
{\tt\small liuheshan666@gmail.com}
\and
Qunshu Lin\\
Zhejiang University\\
{\tt\small linskysuka@gmail.com}
\and
Zachary Bright\\
Exascale Labs\\
{\tt\small zack@exascalelabs.org}
\and
Shitao Tang\\
Simon Fraser University\\
{\tt\small shitaot@sfu.ca}
\and
Yihui He\\
Carnegie Mellon University\\
{\tt\small yihuihe.yh@gmail.com}
\and
Minghao Liu\\
2077AI\\
{\tt\small dreamforever.liu@gmail.com}
\and
Ling Zhu\\
Exascale Labs\\
{\tt\small zhufall@gmail.com}
\and
Cindy Le\\
Columbia University\\
{\tt\small xl2738@columbia.edu}
}
\usepackage{svg}
\begin{document}
\twocolumn[{%
\maketitle
\begin{center}
    \centering
    \includegraphics[width=0.93\linewidth]{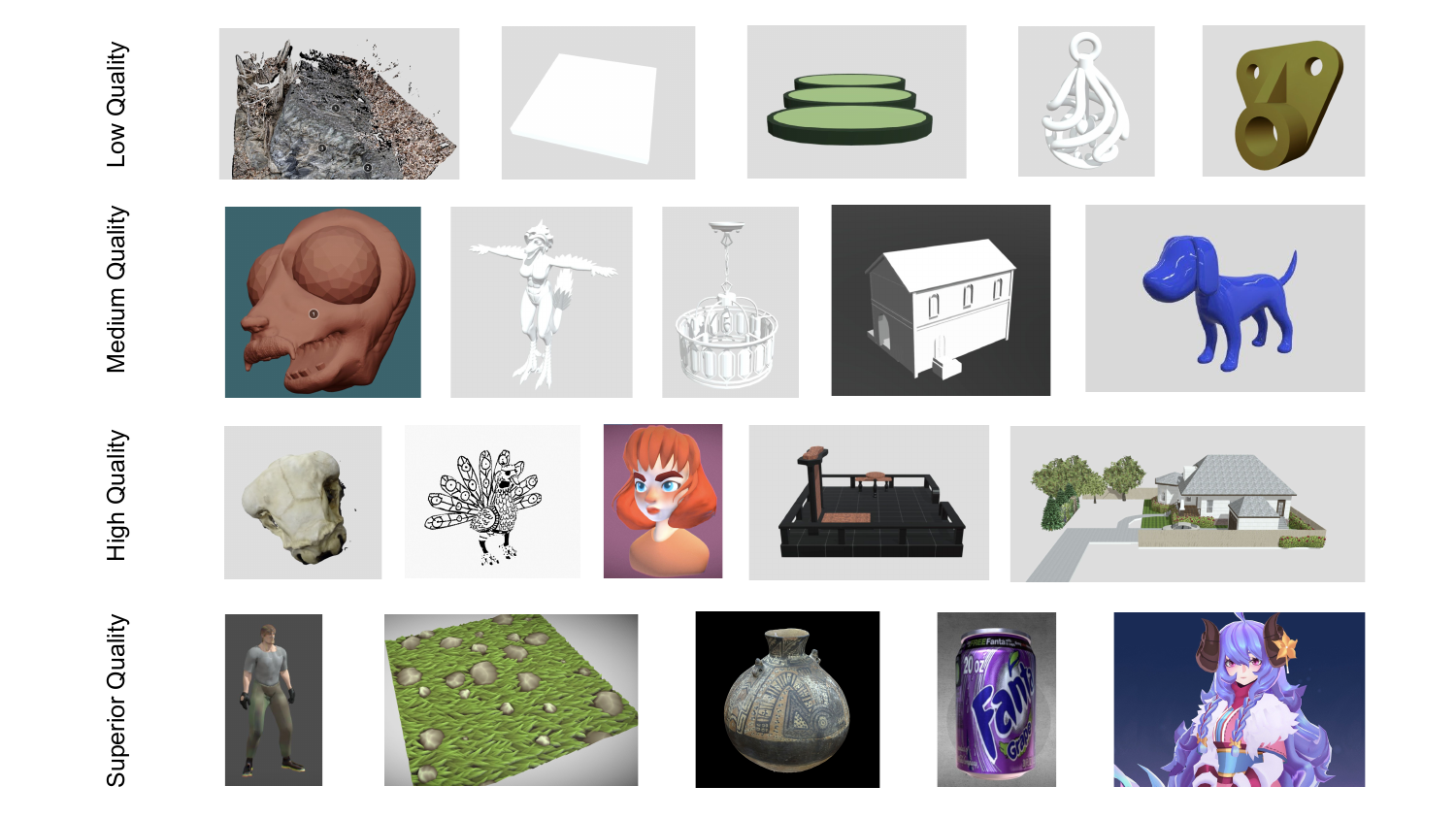}
    \captionof{figure}{Examples of different quality scores assigned to 3D models. (a) \textbf{Low Quality}: Models with no clear semantic meaning due to lack of identifiable structure. (b) \textbf{Medium Quality}: Identifiable objects but lacking essential material texture and color details. (c) \textbf{High Quality}: Models with clear identity and reasonable aesthetic value, featuring basic textures and colors that convey the character of the object. (d) \textbf{Superior Quality}: Professionally textured models with high semantic clarity and aesthetic harmony.} 
    \label{fig:score_examples}
\end{center}
}

]

\begin{abstract}
This paper presents Objaverse++, a curated subset of Objaverse enhanced with detailed attribute annotations. Recent advances in 3D content generation have been driven by large-scale datasets such as Objaverse~\cite{objaverse}, which contains over 800,000 3D objects collected from the Internet. To address the issue of the prevalence of low-quality models in Objaverse, human experts manually annotate 10,000 3D objects with detailed attributes, including aesthetic quality scores, texture color classifications, multi-object composition flags, transparency characteristics, etc. Then, we trained a neural network capable of annotating the tags for the rest of the Objaverse dataset. Through experiments and a user study on generation results, we demonstrate that models pre-trained on our quality-focused subset achieve better performance than those trained on the larger dataset of Objaverse in image-to-3D generation tasks. In addition, by comparing multiple subsets of training data filtered by our tags, our results show that the higher the data quality, the faster the training loss converges. These findings suggest that careful curation and rich annotation can compensate for the raw dataset size, potentially offering a more efficient path to develop 3D generative models. We released our enhanced dataset of approximately 500,000 curated 3D models \footnote{https://github.com/TCXX/ObjaversePlusPlus} 
\footnote{https://huggingface.co/datasets/cindyxl/ObjaversePlusPlus}
to facilitate further research on various downstream tasks in 3D computer vision and aim to extend our annotations to cover the entire Objaverse dataset.

\end{abstract}
\section{Introduction}
\label{sec:intro}

3D modeling is widely used in the media, gaming, and education industries for character and environment creation, realistic product visualizations, animations, and interactive simulations for learning.

Early work applied Generative Adversarial Networks (GAN)~\cite{wu2017learning, chen2018text2shape} and Contrastive Language Image Pre-Training (CLIP)~\cite{jain2022zeroshot,Mohammad_Khalid_2022, michel2021text2mesh} to 3D generation. In recent years, with the accessibility of pre-trained large language models, many researchers have adopted diffusion-based methods for better quality 3D generation ~\cite{le2024euclidreamerfasthighqualitytexturing, poole2022dreamfusion, tang2023mvdiffusionenablingholisticmultiview}.

The emergence of text-to-3D generation models has revolutionized how we create 3D content, yet these models face a critical challenge: They require massive datasets for training, often at the expense of quality control and computational efficiency. Although existing data sets of 3D objects like ShapeNet~\cite{chang2015shapenet} and Objaverse~\cite{objaverse} are useful to the research community, they also have limitations in data quality and lack important information for model training. This challenge affects both research laboratories, where computational resources are often limited, and industry applications, where specific quality standards must be met. This dual perspective drives our investigation into three fundamental questions about 3D datasets and their utilization.

First, can we establish standard rubrics for identifying high-quality 3D objects to support various research tasks? We emphasize critical 3D attributes such as texture quality and transparency, as these features significantly impact model training and generation quality. The distinction between single objects and complex scenes is crucial, as these categories require different generation approaches and algorithmic considerations \cite{yang2024scenecraftlayoutguided3dscene, scenedreamer}. This information will allow us to make informed decisions about data selection for specific tasks while helping industry practitioners identify suitable models for their applications. 

Second, can we automate the annotation of these traits using machine learning? Since ImageNet~\cite{russakovsky2015imagenetlargescalevisual} demonstrated that carefully curated datasets could revolutionize computer vision, dataset curation has played a crucial role in advancing AI capabilities. However, while image quality assessment has matured significantly, automated quality evaluation of 3D content remains an understudied problem. Current approaches often rely on manual curation, making it challenging to scale while maintaining high standards. Inspired by LION5B~\cite{schuhmann2022laion}, we argue that automated quality assessment of 3D models could similarly accelerate progress in the 3D field. The ability to automate the process would not only advance our understanding of 3D aesthetic assessment, but would also provide valuable tools for content filtering and quality control in production environments.

Third, how will high-quality 3D data impact model training in terms of quality and speed? While current state-of-the-art models rely on massive datasets like Objaverse with over 800,000 objects, maintaining and training on such large datasets requires substantial computational resources. This limits accessibility to many researchers and organizations. Furthermore, training on large, unfiltered datasets can introduce noise and inconsistencies that potentially hinder model convergence and final performance. We hypothesize that a carefully curated subset of high-quality 3D models might yield similar or even better results, making advanced 3D generation more accessible to both researchers with limited computational resources and companies seeking cost-effective solutions.

To our knowledge, our work is the first -

\begin{enumerate}

\item to provide quality annotations for objects in the Objaverse dataset and manually annotated the largest scale of 10,000 unique 3D objects;

\item to examine the correlations between the quantity of training data and the quality of generation in the research domain of 3D modeling; and

\item to develop standard rubrics of quality scores and other relevant traits for 3D objects.

\end{enumerate}

We released our dataset 
and hope it can help democratize advanced 3D generation capabilities, make them more accessible across diverse industries, and empower the research community in the field of 3D computer vision.

\section{Related Work}

\textbf{Large-Scale 3D Datasets.}
The emergence of large-scale 3D datasets has been crucial for advancing 3D vision and generation tasks. Notable datasets include ShapeNet~\cite{chang2015shapenet}, which contains approximately 3,000,000 3D models across 3135 object categories. A more recent 3D dataset, Objaverse~\cite{objaverse}, represents a significant milestone in this domain, providing over 800,000 annotated 3D objects sourced from Sketchfab under Creative Commons licenses. A comprehensive collection widely used by the research community, Objaverse still has limitations such as a lack of high-quality and texture-rich objects valuable to train texture generation models. Second, since many objects come from portfolio-sharing platforms by 3D artists, they may have high artistic value and yet not typical and helpful for model training. We believe data quality is as important as quantity and our Objaverse++ will enhance the usability of the existing dataset with additional quality annotations.

\textbf{3D Generation.} Using machine learning to generate 3D models has been an important research question for years. Early 3D generation research was influenced by 2D text-guided generation, starting with GAN-based approaches like 3D-GAN ~\cite{wu2017learning}. After the introduction of Contrastive Language-Image Pre-Training (CLIP)~\cite{radford2021learning} in 2021, several works adapted CLIP-based 2D generation methods to the 3D space~\cite{jain2022zeroshot, Mohammad_Khalid_2022, michel2021text2mesh}. Dreamfields ~\cite{jain2022zeroshot} and DreamFusion ~\cite{poole2022dreamfusion} pioneered a new era of pre-trained text-to-image models for 3D generation through Score Distillation Sampling (SDS), followed by subsequent diffusion-based works ~\cite{lin2023magic3d, chen2023fantasia3d, le2024euclidreamerfasthighqualitytexturing,tang2023mvdiffusionenablingholisticmultiview}. Other works seek improvement through feed-forward methods and train their models based on the Objaverse dataset ~\cite{hong2024lrmlargereconstructionmodel,tochilkin2024triposrfast3dobject}. Image-to-3D generation is a similar research task that builds the 3D structure of an object from a single image or multiple images, which have been addressed by works like ~\cite{Liu_2023_ICCV} and ~\cite{NEURIPS2023_4683beb6}. These methods demonstrate impressive capabilities but require extensive computational resources and massive datasets for training. Our work contributes to this area by investigating whether comparable results can be achieved with a smaller, more carefully curated dataset, potentially making text-to-3D generation more accessible to researchers with limited computational resources.

\textbf{3D Generation Tasks and Dataset Requirements.} Large-scale 3D datasets are utilized in a wide range of 3D tasks. Texture generation is the research task to create color information for a 3D object's surface, usually achieved through 3D generation conditioned by geometric shape of the input object ~\cite{le2024euclidreamerfasthighqualitytexturing, richardson2023texturetextguidedtexturing3d,chen2023text2textextdriventexturesynthesis}. Another common use case, unbounded scene synthesis can be challenging and adopts a system of approaches \cite{yang2024scenecraftlayoutguided3dscene, scenedreamer, he10424939} different than that of single-object generation tasks. Many 3D research tasks rely on large 3D datasets and have diverse needs of training data. Our work proposes and annotates traits that should facilitate the model training for these 3D-related tasks, such as whether a 3D model contains meaningful texture or represents a scene rather than a single object. 

\textbf{3D Dataset Curation and Efficiency.} Machine learning approaches to quality assessment have shown promise across various 3D data domains. In medical imaging, automated assessment of 3D scans has reduced manual inspection workload by 30-70\% while maintaining human-level reliability ~\cite{neuralimaging}. In computer graphics, ~\cite{meshquality} developed deep learning-based quality metrics for textured 3D meshes and annotated variations of 55 source objects, demonstrating the potential of learning-based approaches through extensive human judgment collection. While these studies focus on specific aspects like compression artifacts or medical imagery, the broader challenge of curating large-scale 3D datasets for AI training remains understudied. The closest work to ours is ShapeNet ~\cite{chang2015shapenet}, which provided manually verified category labels for over a million 3D models, though their focus was on semantic categorization rather than quality assessment. Our work bridges this gap by developing systematic quality metrics and demonstrating their impact on model training efficiency.

\section{Objaverse++}
\label{sec:methods}


We manually annotated 10,000 3D objects from Objaverse with quality scores and additional binary traits and trained a neural network capable of annotating the rest of the dataset. 

Our data labeling follows a systematic multistage annotation and validation process.
During the preliminary assessment phase, we consulted artists to determine the essential components of our dataset and randomly sampled 1,000 objects for evaluation. This helps us to understand the data, develop standard labeling rubrics, and find human annotators who can align with our manual labeling results. Building on these initial insights, we develop comprehensive labeling rubrics that standardize the evaluation process. Then, our human annotators manually annotated 10,000 objects with quality and aesthetic tags. Using these data, we trained an annotation network that classifies models and outputs the quality and aesthetic tags based on 2D-rendered multiviews.

\subsection{Annotation Tags}

Our dataset encompasses both quantitative quality scores and qualitative attributes, including style categorization and specific feature tags. During our preliminary assessment, we found that labeling can be aligned across human annotators to a great extent despite the subjective nature of aesthetics and traits. The scoring criteria are carefully calibrated throughout the process to ensure a consistent and meaningful quality assessment.

\subsubsection{Quality and Aesthetics Score}


We define quality score as a metric to tell how useful a 3D object is for machine learning training. We assume a neural network for 3D-related tasks may want to learn two aspects: the semantic meaning of the geometric shape and the color information from the surface texture. The following are the enumeration values for the quality annotation.

\begin{itemize}
\item \textbf{Low Quality}: No semantic meaning. Specifically, if the annotators are not able to identify the object, or if the object is corrupted, it will fall under this category. 

\item \textbf{Medium Quality}: The object is identifiable, but missing the basic material texture and color information. Items that are single-colored by nature, for example, a garden statue, are still considered as having a texture.

\item \textbf{High Quality}: High Quality indicates an acceptable quality with a clear identity of the object. The object is properly textured with some material and color details.

\item \textbf{Superior Quality}: The object is of excellent quality with high semantic clarity. The object is professionally textured with strong aesthetic harmony. This type of object can be used in specific gaming scenario setups without a sense of violation.

\end{itemize}

A rubric and various samples, as shown below and in Fig.\ref{fig:score_examples} are provided to human annotators to determine the score of a 3D model from Objaverse. Note that the quality score is meant to guide researchers and does not necessarily conclude the artistic value of the 3D objects.

Our quality score only has four possible values because the aesthetic judgment of 3D objects is challenging, more subjective than that of 2D images, and requires professional skills in 3D modeling. Previous quality assessment work ~\cite{hethinking} adopts a method in which annotators are asked to rate images in a normal distribution, which is nearly impossible for 3D objects. Fortunately, for a 3D object with a meaningful shape and a colored texture, our annotators only need to decide between the two quality tags, high or superior.

\subsubsection{Binary Traits}





\begin{figure*}[h!]
    \centering
    \includegraphics[width=0.9\linewidth]
                   {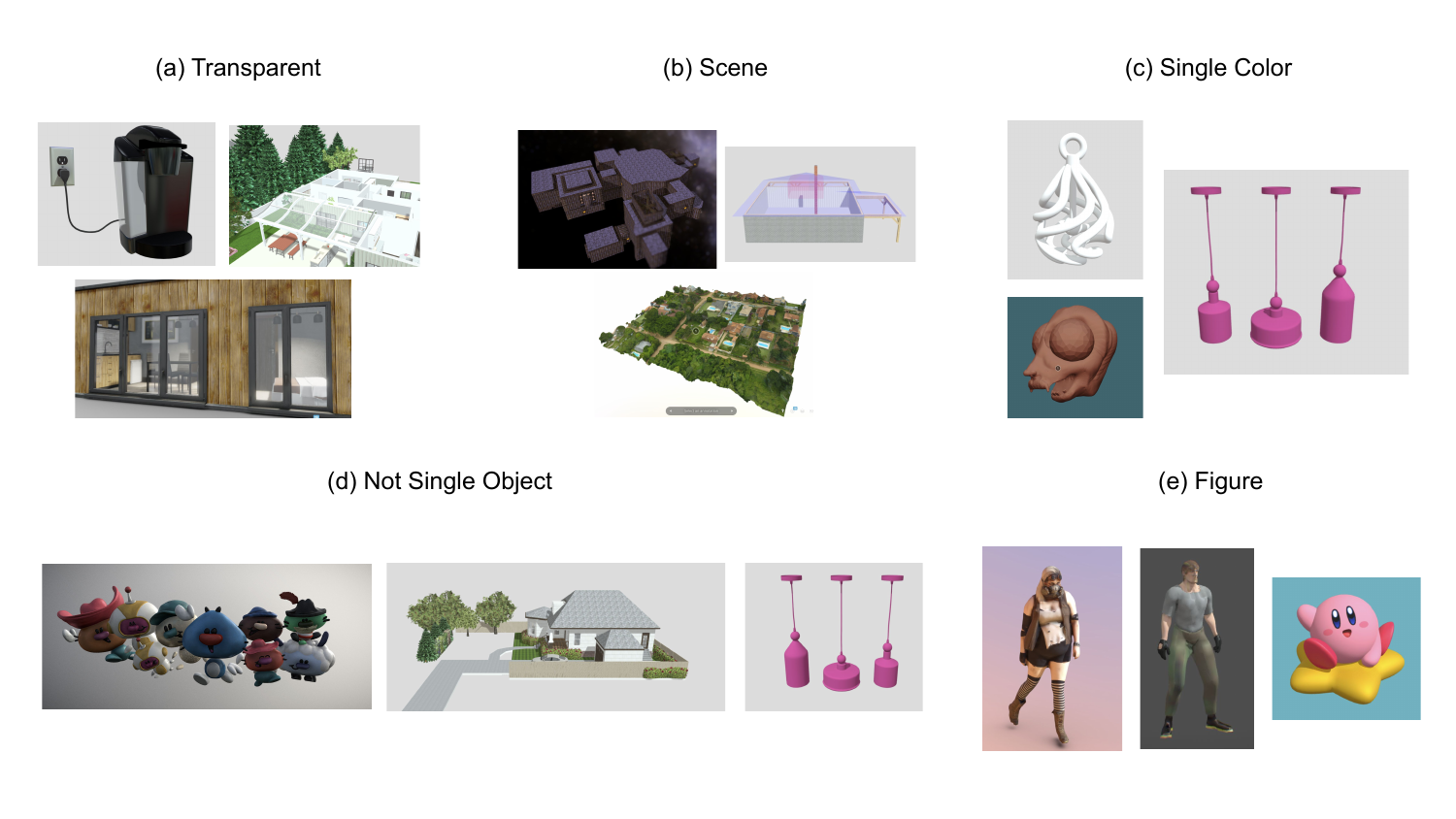}                 
       \caption{Examples of different binary tags assigned to 3D models. (a) \textbf{Transparency}: Identifies models with see-through parts. (b) \textbf{Scene}: Distinguishes scene-like models from standalone objects, enabling differentiation in 3D model generation suited for environments versus single objects. (c) \textbf{Single Color}: Tags unintentionally monochromatic models, filtering out non-texture-rich objects in texture generation learning. (d) \textbf{Not a Single Object}: Identifies models with multiple separate components, focusing learning on single-object generation tasks. (e) \textbf{Figure}: Marks character or figure models, creating a subset for character generation that may benefit from specialized training.} 
    \label{fig:binary_tags_examples}
\end{figure*}

In addition to the quality score on a linear scale, our Objaverse++ annotates several binary tags for each 3D object. The following are all the tags that human annotators are asked to flag out.
\begin{itemize}
\item \textbf{Transparency}: Identifies models with see-through parts, where some areas allow visibility through objects in front of them. Entirely opaque objects do not have this tag. Some 3D generation algorithms rely on 2D multiviews and may not handle transparent parts properly. For example, the results of their generation may project the drivers and the seats inside the car onto the windows of the car. This tag is designed to mitigate such problems.
\item \textbf{Scene}: Identifies whether the model represents a scenario or an environment, rather than a standalone object. Since scene generation \cite{yang2024scenecraftlayoutguided3dscene, scenedreamer} and object generation differ greatly in algorithms and training data needs, this tag will make an important distinction.
\item \textbf{Single Color}: Tags models that are unintentionally monochromatic, meaning that they consist of only one color without any shading, texture, or other visual variation. Models with deliberate monochromatic design (e.g., a sculpture), shading, or texture do not receive this tag. This tag filters out 3D objects that are meaningful for learning texture generation.
\item \textbf{Not a Single Object}: Marks models that consist of multiple separate components rather than a single unified object. The tag refocuses on model learning for the generation and understanding of single objects.
\item \textbf{Figure}: Indicates if the model represents a character, person, or figure. The tag creates a subset of data for potential training in character generation, which may require additional optimization.

\end{itemize}

Researchers can use these tags to filter objects based on certain criteria in various cases. Note that a 3D model could own multiple binary tags. For example, the purple pendant lamp in Fig.\ref{fig:binary_tags_examples} is tagged as both monochromatic and not a single object. Tags are annotated independently and are meant to be orthogonal to each other, but correlation can exist: for example, a 3D model consisting of multiple objects is often a scene, therefore with both the multi-object tag and the scene tag.


\subsection{Labeling Infrastructure}
Using our rubrics, we trained human annotators with a background in 3D model aesthetics to annotate an initial batch of 1000 objects. Their annotation results were then refined and aligned with the established guidelines. These annotators will subsequently annotate a randomly sampled set of 10,000 objects from Objaverse, named Objaverse++ Core, which serves as the training data for the annotation network.

We customized an efficient web-based 3D model annotation platform that enables rapid assessment of multiple quality attributes. The platform features a 3D viewer with intuitive mouse controls for viewing models, with or without edge lines. To optimize annotation efficiency and overcome potential network limitations, we deployed a local caching system for Objaverse models on local servers.

As the annotation platform supports cross-validation and batch management, we divided the data into several batches for better alignment. For each batch of thousands of 3D objects, we sampled a percentage of labeled data and had another set of annotators to validate the annotation results. We addressed discrepancies and aligned with annotators after each batch was cross-validated.

\subsection{Annotation Network}
\label{sec:annotation_network}

The data distribution of binary tags in the human-annotated training dataset is presented in Fig.\ref{fig:binary_tag_distribution}. To scale our approach to cover a larger portion of the Objaverse dataset, we develop and train a 3D model classifier using our manually annotated data.

\subsubsection{Classifier Architecture}

\begin{figure}
    \centering
    \includegraphics[width=1.\linewidth]
    {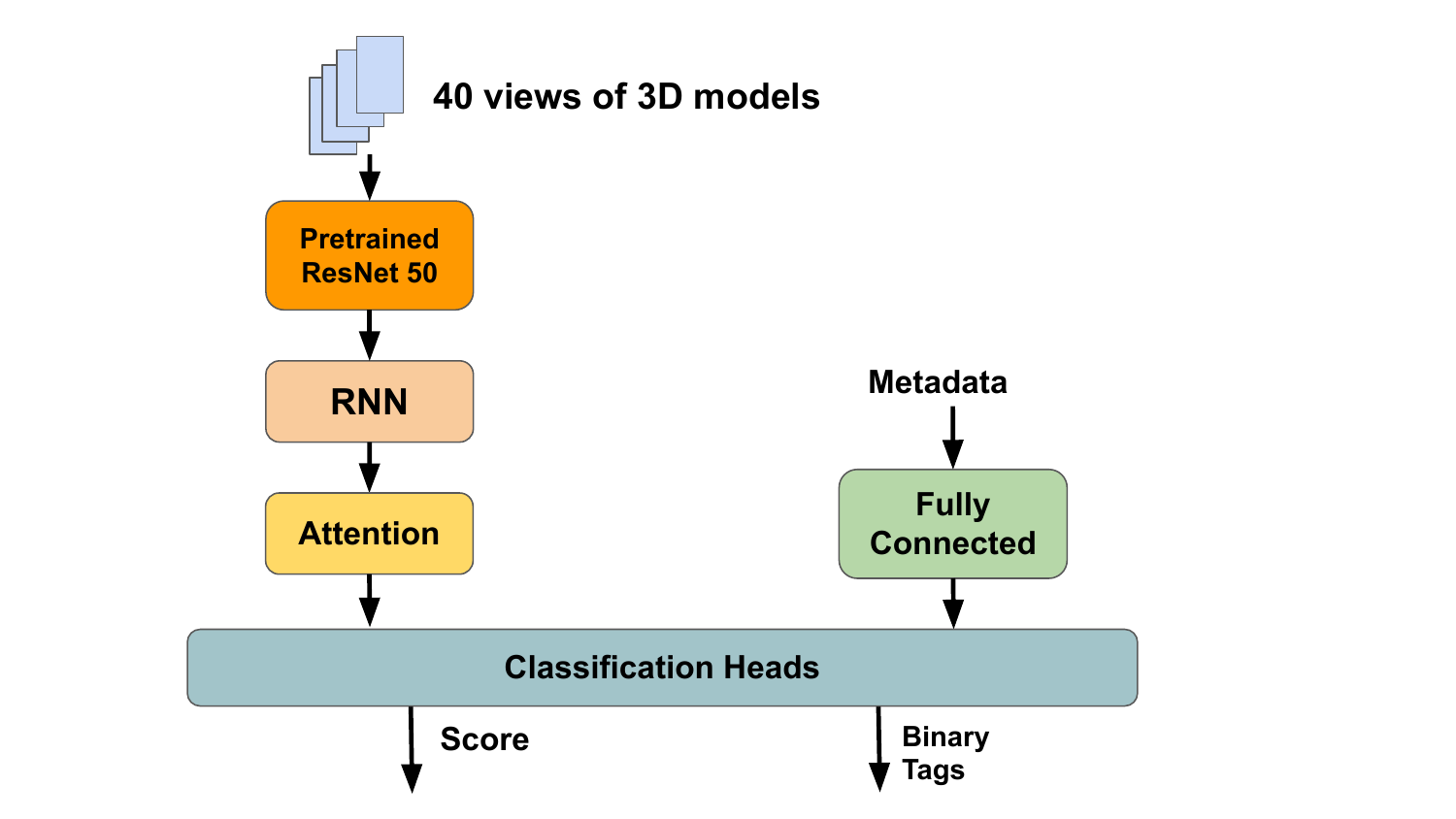}              
   \captionof{figure}{Structure of the annotation network. The network takes 40 different views of 3D models as input, processed by a pre-trained ResNet-50 backbone. Features extracted by the ResNet-50 are passed through an RNN with an attention mechanism to capture spatial dependencies across views. Metadata is fed into a fully connected layer before joining the main pipeline. The combined features are then sent to classification heads for scoring and binary tag predictions.}
   \label{fig:model_structure}
\end{figure}

The 3D object classifier uses a multiview approach, combining convolutional and recurrent neural networks with an attention mechanism, enhanced by object-specific metadata. The training data includes 40 screenshots of each 3D model, captured from different angles, along with metadata from Objaverse. This input format, commonly used in works such as \cite{su2015multi} and \cite{yu2018multi}, is both intuitive and straightforward to implement. However, it has limitations, particularly in distinguishing density without wireframe visualization, but it aligns closely with how humans visually assess and evaluate 3D models. The model architecture is shown in Fig.\ref{fig:model_structure} and has five main components:
\begin{itemize}
    \item \textbf{Feature Extraction}: A pre-trained ResNet50 \cite{he2016deep} is used as our feature extractor to process each 2D projection of the 3D object independently. Given a set of 40 random views of a 3D model, ResNet50 produces corresponding feature vectors that capture visual information from each perspective.
    
    \item \textbf{Sequence Modeling} with Recurrent Neural Network (RNN): The sequence of feature vectors is fed into a Long Short-Term Memory (LSTM) network that models dependencies across views. The LSTM generates a sequence of hidden states encoding information about the structure of the object as observed from each view.
    
    \item \textbf{Attention Mechanism}: To focus on the most informative views, an attention layer is applied over the LSTM hidden states. The attention mechanism computes weights for each hidden state of the view, resulting in a context vector. This context vector emphasizes features of views that are deemed critical for classification.

    \item \textbf{Metadata Integration}: In addition to visual features, we incorporate metadata such as vertex and edge counts, view counts, and like counts, and process them by a fully connected layer. The resulting metadata embedding is concatenated with the attention-weighted feature vector, providing additional context for classification.

    \item \textbf{Classification Heads}: The combined representation is passed through multiple classification heads, each predicting a specific attribute of the 3D model, such as style, quality score, and binary tags.

\begin{figure}
    \centering
    \includegraphics[width=0.92\linewidth]
                   {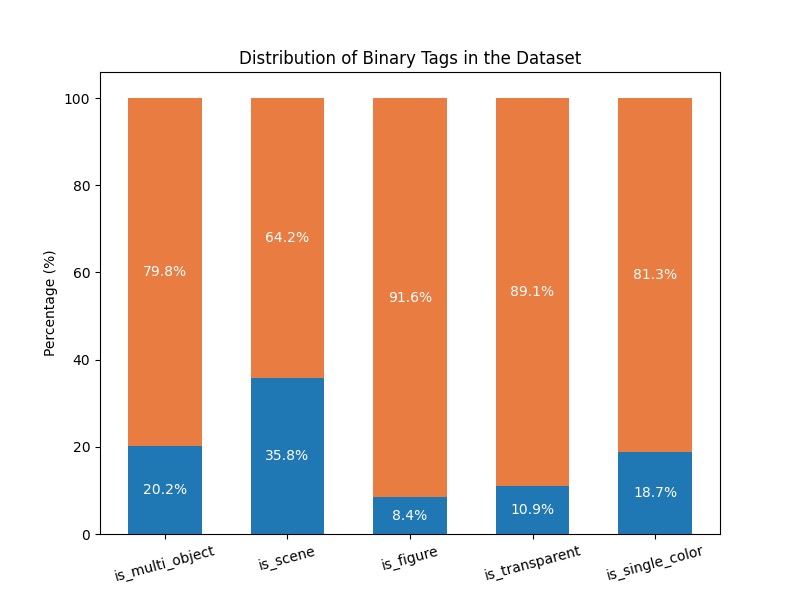}                 
   \captionof{figure}{The distribution of binary tags in the human-annotated training dataset. }
   \label{fig:binary_tag_distribution}
\end{figure}

\end{itemize}

Cross-entropy loss is used for score annotation, and BCEWithLogitsLoss is used for binary tag labeling.

Among the many existing embedding methods for 3D objects~\cite{sitzmann2019deepvoxelslearningpersistent3d, uy2020deformationaware3dmodelembedding, liu2024isotropic3dimageto3dgenerationbased}, we adopted the current embedding approach based on 2D multiview, whereas an alternative solution involves reading 3D geometric and textural information. We acknowledge that using 3D-native embedding would unlock the potential of labeling certain tags. For example, a watertightness annotation can be useful and is achievable only through 3D-native embeddings. Similarly, annotation on mesh density is beneficial to the 3D modeling industry as objects with different levels of polygon counts are adopted in different use cases. Due to its prevalence and easy access, we explored more with a 2D-based method for our annotation network.

\subsubsection{Classifier Validation}

The evaluation metrics performed on the test set (1971 samples) are shown in Table \ref{tab:annotation_metrics}.

\begin{table}[h!]
    \centering
    \caption{Metrics of Annotation Network}
    \label{tab:annotation_metrics}
    \begin{tabular}{lccc}
        \hline
        \textbf{Metric} & \textbf{Accuracy} & \textbf{F1 Score} & \textbf{mAP}\\
        \hline
        score* & 0.5945 & --  &  --\\
        relaxed score accuracy & 0.8221 & -- & --\\
        is\_multi\_object & 0.8621 & 0.703 & 0.6927\\
        is\_scene & 0.8667 & 0.731  & 0.9176\\
        is\_figure & 0.9448 & 0.844 & 0.6815\\
        is\_transparent & 0.9372 & 0.835& 0.7435 \\
        is\_single\_color & 0.9169 & 0.796 & 0.6745\\
        \hline
    \end{tabular}
\end{table}

Among all the metrics, it is clear that the metrics for the binary tags (is\_multiple\_object, is\_scene, is\_figure, is\_transparent) demonstrate strong accuracies. These metrics highlight the reliability of the network in identifying different characteristics of the model, which is valuable for various 3D modeling applications. ``Score" (marked with * in Table \ref{tab:annotation_metrics}) is a relatively weak metric of the annotation network due to subjective nuances, as mentioned above. Here, we include a ``relaxed accuracy for score" that allows scores 2 and 3 to be interchangeable. The ``relaxed score accuracy" improves significantly to 82.21\%, showing that the network captures general quality distinctions effectively, though missing subtle differences among models of high quality.

The high accuracy of our annotation network shows that our proposed tags are learnable by a carefully designed classifier. 
\section{Dataset Evaluation}

To evaluate the dataset, we set up an image-to-3D generation task as a practical and reproducible approach for future studies on 3D dataset curation. OpenLRM \cite{openlrm}, an open source framework designed to generate 3D models from a single image input. It allows for fine-tuning as well as training from scratch, making it suitable for both adapting pre-trained models and building customized ones tailored to specific datasets. Given our computational constraints, we utilized OpenLRM’s small model architecture, which is optimized for limited computing resources while still providing effective 3D generation capabilities. Table \ref{tab:model_configuration} describes the model configuration used in this work\cite{openlrm}.

\begin{table*}[h!]
    \centering
    \small
    \begin{tabular}{c|c|c|c|c|c|c|c}
        \hline
        \textbf{Type} & \textbf{Layers} & \textbf{Feat. Dim.} & \textbf{Attn. Heads} & \textbf{Triplane Dim.} & \textbf{Input Res.} & \textbf{Image Encoder} & \textbf{Size} \\
        \hline
        small & 12 & 512 & 8 & 32 & 224 & dinov2\_vits14\_reg & 446M \\
        \hline
    \end{tabular}%
    \caption{OpenLRM model configuration details.}
    \label{tab:model_configuration}
\end{table*}

We randomly sampled 100,000 objects from Objaverse to create Training Set A, and its binary quality tag distribution is described in Table \ref{tab:binary_tag_distribution}. Using quality filtering criteria, we then selected approximately 50,000 high-quality objects to form Training Set B. Filtering criteria included selecting models of high or superior quality, excluding monochromatic models, excluding scenes, and excluding models with any transparent part. This setup is well-suited for single-object generation, as the filtering criteria ensure a subset that closely aligns with the input distribution of the generation task.

\begin{table*}[htbp!]
    \centering
        \begin{tabular}{l|c|c|c|c|c|c}
            \hline
            \textbf{Label} & \textbf{is\_multi\_object} & \textbf{is\_scene} & \textbf{is\_figure} & \textbf{is\_transparent} & \textbf{is\_single\_color} \\
            \hline
            0 (No) & 94.98\% & 59.45\% & 97.64\% & 97.67\% & 81.32\% \\
            1 (Yes) & 5.02\% & 40.55\% & 2.36\% & 2.33\% & 18.68\% \\
            \hline
        \end{tabular}
    \caption{Distribution of Selected Binary Tags in the 100,000 Annotation Dataset.}
    \label{tab:binary_tag_distribution}
\end{table*}

We trained the same model on Training Set A. This training, starting from scratch, took approximately 9 hours on 8 H100 GPUs. Subsequently, we trained the same model on Training Set B, which required about 6 hours under the same conditions.

\begin{figure}
    \centering
    \includegraphics[width=0.4 \textwidth]{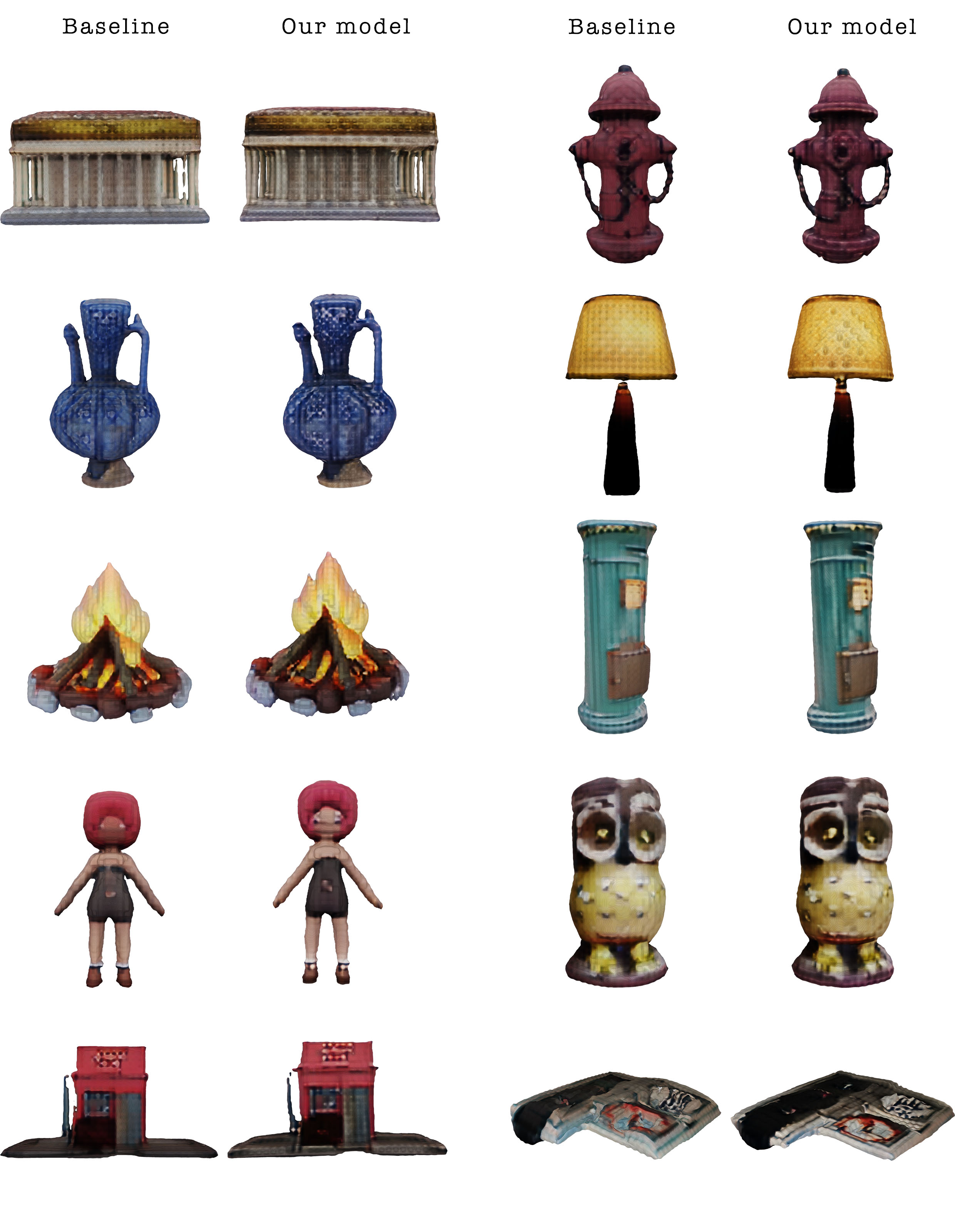}
    \caption{A comparison of image-to-3D generation results by a randomly sampled 100,000 dataset vs. our model. } 
    \label{fig:results}
\end{figure}

To assess the performance of the model, we conducted a user study with 47 participants. Each participant was presented with 10 pairs of generation results: one generated from Training Set A and the other from Training Set B. The generated 3D models are inferenced given the sample inputs of the open-source OpenLRM. Participants were asked to choose their preferred result or to select ``no preference" without knowing the origin of each generation. To reduce bias, the question order and the option order were randomized. The participants came from diverse backgrounds, including artists, machine learning researchers, game developers, and software engineers. 

\begin{figure}
    \centering
    \includegraphics[width=0.4 \textwidth]{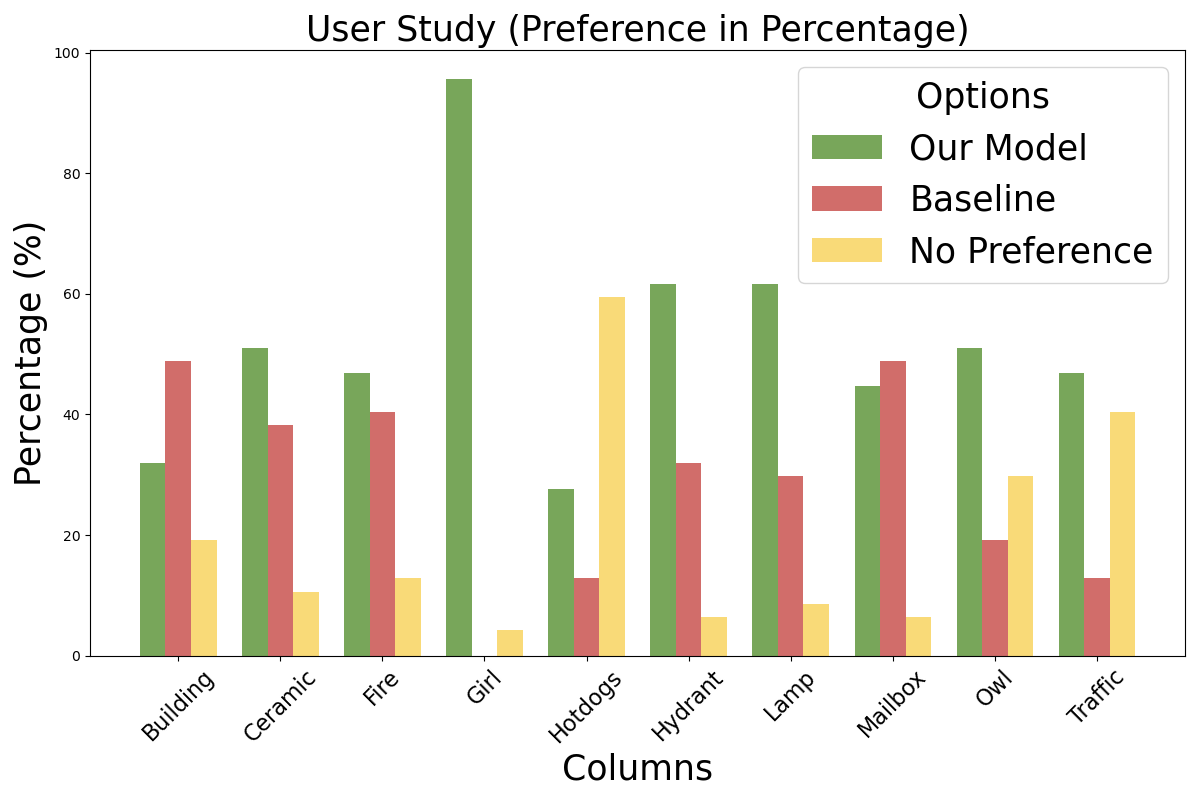}
    \caption{User Study Results. Of the 10 pairs of objects, 8 showed a preference for our model over the baseline. For pairs like the one titled ``girl," more than 95\% of the participants chose our result. For ``mailbox", ``hydrant" and ``lamp", despite the presence of the no preference option, the majority of the participants chose our generation, proving the quality is significantly higher than baseline. } 
    \label{fig:user-study}
\end{figure}

 \begin{figure}
    \centering
    \includegraphics[width=0.43 \textwidth]{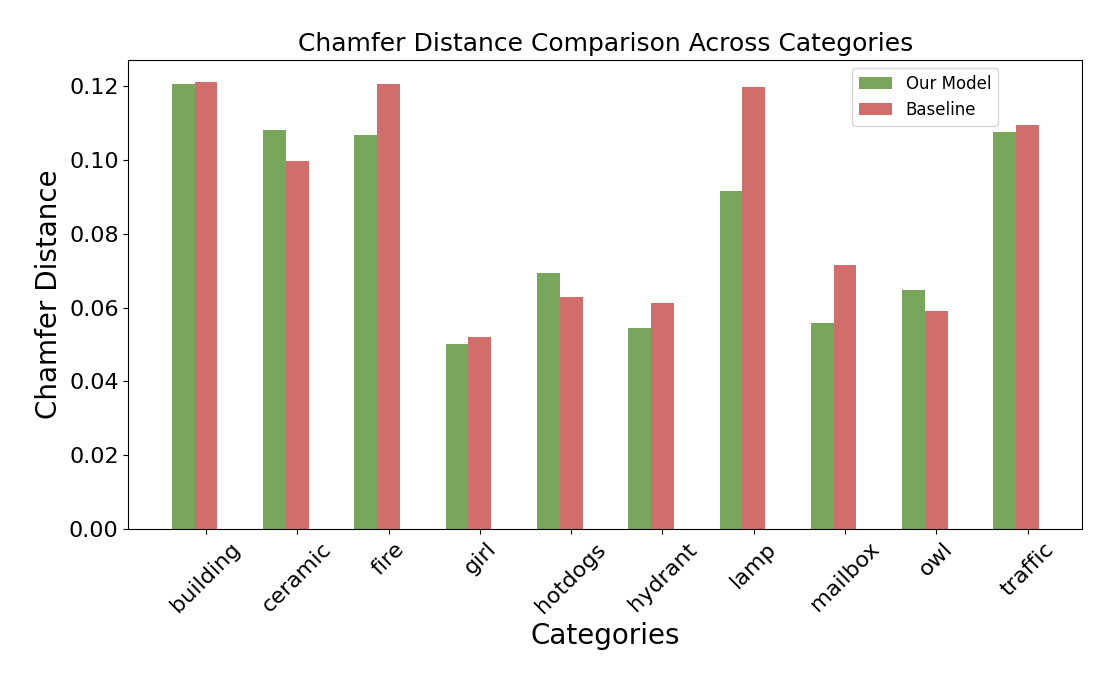}
    \caption{Chamfer distance comparison of the model trained using the randomly-sampled 100,000 dataset vs. our model using high-quality subset. The smaller chamfer distance indicates a more similar generated model to the ground-truth 3D model. Our model outperforms the baseline in 7 out of 10 experiments, which aligns with the user study result.} 
    \label{fig:chamfer}
\end{figure}

\subsection{Better Generation Quality}

The user study shows significant results (see Fig.\ref{fig:user-study}) in favor of our generation model. Of the 10 pairs of objects, 8 showed a preference for our model over the baseline. Despite the presence of the no-preference option, participants favor our generation more. Our model receives 83.5\% more votes than the baseline.

To further prove the significance of our results, we have conducted quantitative research using chamfer distance. The pre-trained OpenLRM-obj-small-1.1 trained using the whole Objaverse is used as the ground truth, and we compare the performance of our model (high-quality subset) and the
baseline model (randomly sampled 100k subset) as shown in Fig.\ref{fig:chamfer}. A lower Chamfer distance indicates a closer match between the generated model and the ground-truth 3D model. Our model achieves better performance in 7 out of 10 experiments, consistent with the findings of the user study.

Besides quantitative measurements, our team and some participants noticed that our model generates 3D objects with higher color contrast. Vibrant color benefits the 3D modeling process and is expected from a refined dataset. As estimated by our annotation results, about one-fifth of the original Objaverse dataset are scanned objects that tend to have high polygon counts and low texture quality, generative models trained on these data will be impacted by these traits. 3D objects with lower polygon counts and vivid color tones resemble those of human artists. 

\begin{figure}
    \centering
    \includegraphics[width=0.95\linewidth]{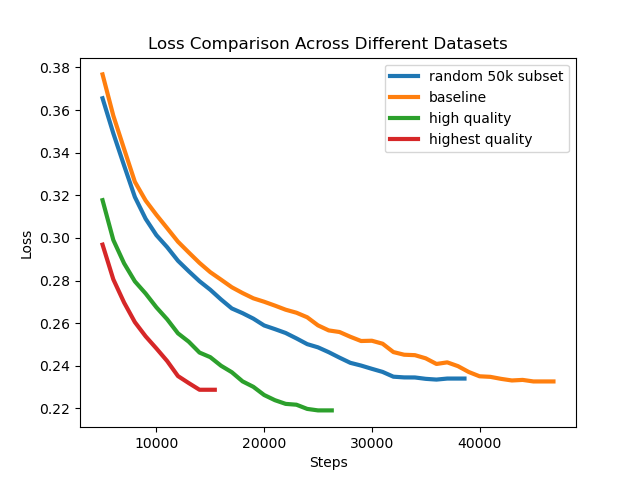}
    \caption{
        \textbf{Baseline}: A randomly sampled subset of 100,000 objects from Objaverse, called Training Set A. It is used as the baseline for comparison. \\
        \textbf{Random 50k Subset}: A randomly sampled subset of 50,000 samples from Training Set A, used to isolate the effect of dataset size on convergence. \\
        \textbf{High-Quality Subset}: A quality-filtered dataset containing samples with high and superior quality, excluding monochromatic models, scenes, and models with any transparent part, curated to improve data quality while maintaining a reasonable sample size, called Training Set B. \\
        \textbf{Superior Quality Subset}: A highly curated subset containing only samples with superior quality, excluding monochromatic models, scenes, and models with any transparent part, aimed at maximizing data quality for optimal model convergence.
    }
    \label{fig:openlrm-loss-convergence}
\end{figure}

\subsection{Faster Convergence in Training}

As shown in Fig.~\ref{fig:openlrm-loss-convergence}, our model on a carefully curated dataset demonstrates faster convergence than a randomly selected subset of the Objaverse. As discussed above, our curation process significantly reduces the number of noisy or low-quality samples. Training in a refined selection allows the model to require fewer epochs and steps to achieve optimal performance.

\noindent \textbf{Quality Over Random Sampling.} To ensure that dataset size is not the primary reason for faster convergence, we randomly sampled 50,000 objects from Training Set A to create a subset of comparable size to our ``high-quality dataset".  As illustrated in Fig.\ref{fig:openlrm-loss-convergence}, the random 50k subset does not result in a significantly faster convergence than the baseline dataset. This implies that simply reducing the dataset size does not guarantee faster convergence. In comparison, our ``high-quality dataset", similar in size to the random 50k subset, produces substantially faster convergence, demonstrating that the curated quality of the samples is crucial.

\noindent \textbf{Impact of Annotation Quality.} The ``superior quality subset", containing only top quality samples, converges faster than the ``high quality subset" dataset, which includes both high and superior quality models. This finding supports the idea that our scoring rubrics are efficient, as higher-quality data directly contributes to faster and more efficient model training.

In summary, the findings show that faster convergence is driven by the quality of the curated dataset rather than its size. This underscores the importance of our quality-filtered data and validates that our labeling process improves the quality of the original Objaverse dataset.

\section{Conclusion}
We used a combination of human labeling and an annotation network to tag models from Objaverse. The tagging results for the 100,000 models sampled from Objaverse will be open-sourced as the initial result, allowing users to create custom training sets tailored to their specific needs. Using quantitative metrics and a user study, we show how high-quality training data can simultaneously enhance effectiveness, efficiency, and performance in 3D generation.

\noindent \textbf{Future directions.} We acknowledge that further experiments comparing Objaverse++ with Objaverse-XL are important.
In addition, we will explore comprehensive and quantitative filtering criteria, possibly leveraging more advanced annotation networks to tag additional model attributes, such as structural complexity or aesthetic style.

\section{Acknowledgement}

We gratefully acknowledge Exascale Labs and Zillion Network for providing the computational resources and supporting our training infrastructure that made this research possible. We thank Abaka AI for their valuable assistance with data labeling. Special thanks to Ang Cao and Liam Fang for their technical and artistic insights that significantly enhanced our understanding of 3D model quality assessment. 

{
    \small
    \bibliographystyle{IEEEtran}
    \bibliography{main}
}

\clearpage
\setcounter{page}{1}
\maketitlesupplementary

\section{Quality Score Rubrics}

\begin{figure}
    \centering
    \includegraphics[width=0.45 \textwidth]{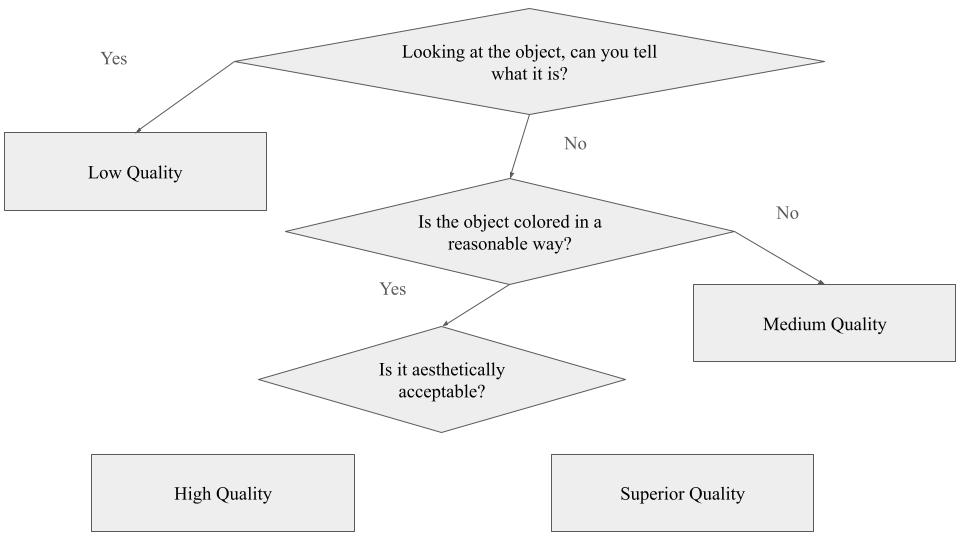}
    \caption{Decision tree for human annotators to categorize the quality level of a 3D object. } 
    \label{fig:rubric_details}
\end{figure}

Due to the specialty of judging 3D objects, we composed additional training material for our human annotators, including Figure~\ref{fig:rubric_details} and the rubrics below. For quality score, here are some criteria to consider when an object has proper semantic meaning and texture.

\textbf{High-Quality Criteria}:
\begin{itemize}

\item Basic color scheme present but lacks richness and aesthetic appeal.
\item Acceptable geometric shapes - not too rough but not highly detailed.
\item Basic textures present - goes beyond flat surfaces but lacks sophistication.
\item Visually comfortable and harmonious, but lacks refinement in details (like color rendering and fabric textures).

\end{itemize}

\textbf{Superior-Quality Criteria}:

\begin{itemize}

\item High-quality modeling with rich textures, vibrant colors, and aesthetic value.
\item Rich, harmonious color combinations that feel natural or appropriate to the style.
\item Geometric proportions that match either real-world references or suit the intended artistic style.
\item Detailed surface texturing with effective lighting/shading.
\item Aesthetically pleasing or visually impactful.
\item Abundant detailed elements such as decorations, patterns, etc.

\end{itemize}

\section{Training Loss}

\begin{figure}
\centering
\includegraphics[width=\linewidth]{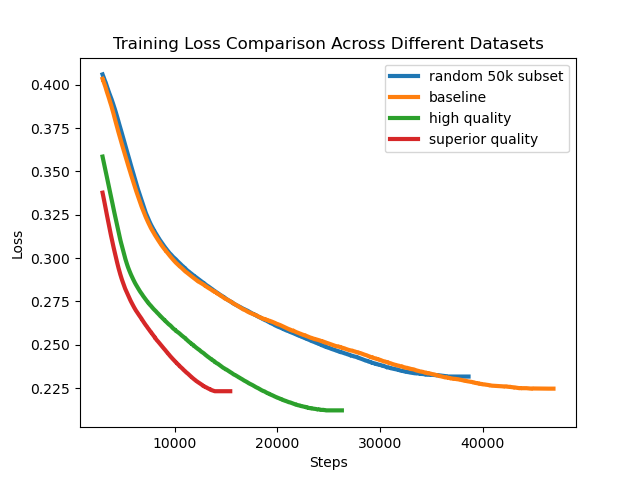}
\caption{
        Training loss comparison across different datasets. Similar to the validation loss result~\ref{fig:openlrm-loss-convergence}, the model converges significantly faster on high-quality and superior-quality datasets, and converges roughly at the same speed on a random 50k subset and baseline, which is a randomly sampled subset of 100,000 objects from Objaverse.
    }
\label{fig:openlrm-train-loss-convergence}
\end{figure}
Figure~\ref{fig:openlrm-train-loss-convergence} demonstrates the impact of dataset quality on training loss. The high and superior quality subsets show faster and more stable convergence than the baseline, a randomly sampled subset of 100,000 objects from Objaverse, and random 50k subsets. Quality-filtered data reduces noise, accelerates optimization, and enhances learning stability, allowing the model to converge more efficiently. In contrast, the baseline dataset’s noisy samples hinder optimization. The superior quality subset achieves the best results among the four datasets, underscoring the importance of high-quality data over dataset size for efficient model training.
\section{User Study Results}

We include an additional Table ~\ref{tab:user_study_percentage} for the percentage breakdown of our user study.


\begin{table}[]
\small
\begin{tabular}{l|r|r|r}
                  & \multicolumn{1}{l}{\textbf{Our model}} & \multicolumn{1}{l}{\textbf{Baseline}} & \multicolumn{1}{l}{\textbf{No preference}} \\
\hline
\textbf{Building} & 31.9                                   & \textbf{48.9}                         & 19.2                                       \\
\textbf{Ceramic}  & \textbf{51.1}                          & 38.3                                  & 10.6                                       \\
\textbf{Fire}     & \textbf{46.8}                          & 40.4                                  & 12.8                                       \\
\textbf{Girl}     & \textbf{95.7}                          & 0                                     & 4.3                                        \\
\textbf{Hotdogs}  & \textbf{27.7}                          & 12.8                                  & 59.5                                       \\
\textbf{Hydrant}  & \textbf{61.7}                          & 31.9                                  & 6.4                                        \\
\textbf{Lamp}     & \textbf{61.7}                          & 29.8                                  & 8.5                                        \\
\textbf{Mailbox}  & 44.7                                   & \textbf{48.9}                         & 6.4                                        \\
\textbf{Owl}      & \textbf{51.1}                          & 19.1                                  & 29.8                                       \\
\textbf{Traffic}  & \textbf{46.8}                          & 12.8                                  & 40.4    
\end{tabular}
\caption{User study results in percentage. Of the 10 pairs of objects, 8 preferred our model (in bold) over the baseline.}
\centering
\label{tab:user_study_percentage}
\end{table}

\end{document}